**Anthony Sigogne, Matthieu Constant, Éric Laporte**

**Université Paris-Est, LIGM**

**5 boulevard Descartes, Champs sur Marne 77454 CEDEX 2 France**

sigogne@univ-mlv.fr, mconstan@univ-mlv.fr, laporte@univ-mlv.fr

# INTÉGRATION DES DONNÉES D'UN LEXIQUE SYNTAXIQUE DANS UN ANALYSEUR SYNTAXIQUE PROBABILISTE

*Résumé :* Cet article présente les résultats d'une évaluation sur l'intégration des données issues d'un lexique syntaxique, le Lexique-Grammaire, dans un analyseur syntaxique. Nous montrons qu'en modifiant le jeu d'étiquettes des verbes et des noms prédicatifs, un analyseur syntaxique probabiliste non lexicalisé obtient des performances accrues sur le français.
*Mots clés :* Analyse syntaxique probabiliste, Lexique syntaxique, Lexique-Grammaire

## 1. INTRODUCTION

Les lexiques syntaxiques sont des ressources très riches pour les langues qui en disposent. Ils contiennent de nombreuses informations utiles comme les cadres de sous-catégorisation qui nous renseignent sur le comportement syntaxique des entrées du lexique. La plupart du temps, ces lexiques concernent uniquement les verbes mais on peut en trouver certains, comme le Lexique-Grammaire (Gross, 1994), qui traitent d'autres catégories grammaticales comme les noms, adjectifs ou encore les adverbes. De nombreux travaux basés sur l'analyse syntaxique symbolique font état de l'utilisation d'un lexique syntaxique, par exemple (Sagot, 2006; Sagot & Tolone, 2009; de La Clergerie, 2010). En ce qui concerne l'analyse syntaxique statistique, on trouve un certain nombre de travaux qui expérimentent l'intégration des données d'un lexique syntaxique dans des analyseurs basés sur des grammaires probabilistes telles que les Grammaires Syntagmatiques guidées par les Têtes [HPSG] (Carroll & Fang, 2004), les Grammaires Lexicales-Fonctionnelles [LFG] (O'Donovan et al., 2005; Schluter & Genabith, 2008) ou encore les Grammaires Probabilistes non-contextuelles [PCFG] (Briscoe & Carroll, 1997; Deoskar, 2008) . Ces derniers ont incorporé des informations de valence au niveau du lexique et de la grammaire et ont observé un léger gain de performance. Cependant, leurs ressources lexicales ont été obtenues automatiquement à partir d'un corpus. De plus, les informations de valence concernaient principalement les verbes. Dans cet article, nous allons voir comment nous pouvons exploiter les données du Lexique-Grammaire afin d'améliorer un analyseur syntaxique probabiliste basé sur une grammaire PCFG.

Nous décrivons, section 2, l'analyseur syntaxique probabiliste utilisé dans le cadre de nos expériences. Dans la section 3, nous présentons succinctement le Lexique-Grammaire ainsi que la version au format Lglex. Nous détaillons les informations présentes dans ce lexique qui peuvent être utilisées dans le cadre de l'analyse syntaxique probabiliste. Ensuite, dans la section 4, nous présentons une méthode d'intégration de ces informations dans le processus d'analyse syntaxique, puis, à la section 5, nous décrivons les expériences et nous discutons des résultats. Enfin, section 6, nous concluons.

## 2. ANALYSE SYNTAXIQUE NON LEXICALISÉ

L'analyseur syntaxique probabiliste, utilisé pour nos expériences, est le Berkeley Parser (appelé BKY par la suite) (Petrov et al., 2006) [1]. Cet analyseur est basé sur un modèle PCFG non lexicalisé. Le principal problème des grammaires hors-contexte non lexicalisées est que les symboles pré-terminaux (étiquettes morpho-syntaxiques) encodent des informations trop générales qui discriminent peu les ambiguïtés syntaxiques. L'avantage de BKY est qu'il tente de remédier au problème en générant une grammaire ayant des symboles pré-terminaux complexes. Il suit le principe des annotations latentes introduites par (Matsuzaki et al., 2005) . Cela consiste à créer itérativement plusieurs grammaires, qui possèdent un jeu de symboles pré-terminaux de plus en plus complexes. A chaque itération, un symbole de la grammaire est séparé en plusieurs symboles selon les différents comportements syntaxiques du symbole qui apparaissent dans un corpus arboré. Les paramètres de la grammaire latente sont estimés à l'aide d'un algorithme basé sur Espérance-Maximisation (EM).

---

[1] L'analyseur BKY est disponible librement à l'adresse http ://code.google.com/p/berkeleyparser/

Dans le cadre du français, (Seddah et al., 2009) ont montré que BKY donne des performances au niveau de l'*état de l'art*. Ils ont également montré que certains analyseurs, basés sur le paradigme lexicalisé (les noeuds syntagmatiques sont annotés avec le mot tête), étaient moins performants que BKY.

## 3. LEXIQUE-GRAMMAIRE

Les tables du Lexique-Grammaire constituent aujourd'hui une des principales sources d'informations lexicales syntaxiques pour le français[2]. Leur développement a démarré dès les années 1970 par Maurice Gross et son équipe (Gross, 1994). Ces informations se présentent sous la forme de tables. Chaque table regroupe les éléments d'une catégorie donnée partageant certaines propriétés définitoires, qui relèvent généralement de la sous-catégorisation. Ces éléments forment une classe. Il existe des tables pour différentes catégories grammaticales (noms, verbes, adverbes,...). Ces tables ont récemment été rendues cohérentes et explicites dans le cadre du travail de (Tolone, 2011)[3], notamment au moyen d'une table des classes. Cette table particulière encode les propriétés définitoires qui sont communes à toutes les entrées d'une classe. Ces propriétés n'étaient présentes initialement que dans la littérature. Par exemple, cette table nous indique que les verbes de la table *V_35LR* acceptent une construction syntaxique de type *N0 V N1*. Cependant, les tables du Lexique-Grammaire ne sont pas directement exploitables par la machine. Nous utilisons donc le format Lglex (Constant & Tolone, 2008), qui est une version structurée des tables au format XML. Chaque entrée des tables du Lglex contient différentes informations telles que le numéro de la table, les arguments possibles ainsi que leur nombre et les constructions syntaxiques acceptées.

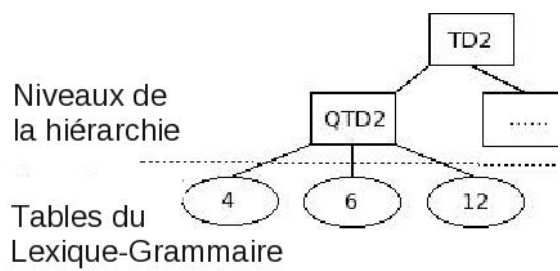

FIG. 1: Extrait de la hiérarchie des tables des verbes

Nous disposons, pour les verbes uniquement, d'une hiérarchie des tables sur plusieurs niveaux[4]. Chaque niveau contient plusieurs classes qui regroupent des tables du Lexique-Grammaire qui ne partagent pas forcément toutes leurs propriétés définitoires mais qui ont un comportement syntaxique relativement similaire. La figure 1 montre un extrait de la hiérarchie. Les tables *4, 6* et *12* sont regroupées dans une classe *QTD2* (transitifs directs à deux arguments avec un objet pouvant être sous la forme d'une complétive). Puis cette classe est elle-même regroupée avec d'autres classes au niveau supérieur de la hiérarchie pour former une classe *TD2* (transitifs directs à deux arguments).

Les caractéristiques de chaque niveau sont indiquées dans le tableau 1[5] (le niveau 0 représente l'ensemble des tables présentes dans le Lexique-Grammaire). On peut noter que les tables des verbes contiennent 5923 formes verbales distinctes pour 13862 entrées différentes. La colonne *#classes* précise le nombre de classes distinctes. Quant aux colonnes *AVG_1* et *AVG_2*, elles indiquent le nombre moyen d'entrées par classe et le nombre moyen de classes par forme verbale distincte.

---

[2] On pourra citer également les lexiques LVF (Dubois & Dubois-Charlier, 1997), Dicovalence (Eynde & Piet, 2003) et Lefff (Sagot, 2010).

[3] Ces ressources sont librement disponibles à l'adresse
http ://infolingu.univ-mlv.fr>Données_Linguistiques>Lexique_Grammaire> Téléchargement

[4] La hiérarchie des tables des verbes est disponible à l'adresse suivante : http ://igm.univ-mlv.fr/~sigogne/arbre-tables.xlsx

[5] On peut également préciser que 3121 verbes (3195 entrées) sont dit non ambigus. C'est à dire que toutes leurs entrées sont présentes dans une seule et même table.

| Niveau | #classes | AVG_1 | AVG_2 |
|---|---|---|---|
| 0 | 67 | 207 | 2.15 |
| 1 | 13 | 1066 | 1.82 |
| 2 | 10 | 1386 | 1.75 |
| 3 | 4 | 3465 | 1.44 |

TAB. 1: Caractéristiques de la hiérarchie des tables des verbes

Le principal avantage d'avoir une hiérarchie des tables est d'obtenir une réduction du nombre de classes associées à chaque verbe présent dans les tables. Nous verrons que la réduction des ambiguïtés est cruciale lors de nos expériences.

**4. EXPLOITATION DES DONNÉES DU LEXIQUE-GRAMMAIRE**

De nombreuses expériences d'analyse syntaxique faites sur le français (Crabbé & Candito, 2008; Seddah et al., 2009), ont montré qu'en raffinant les étiquettes morpho-syntaxiques du corpus d'apprentissage, les performances étaient accrues. Nous allons nous inspirer de ces travaux en intégrant aux étiquettes des informations issues du Lexique-Grammaire. Dans le cadre de cet article, nous utiliserons uniquement les tables des verbes et des noms prédicatifs.

Les numéros de table des entrées du lexique sont des indices importants sur les comportements syntaxiques. Dans une première expérience, appelée *AnnotTable*, nous avons ajouté l'information du ou des numéros de table associés au nom ou au verbe. Par exemple, le verbe *chérir* appartient à la table *12*. L'étiquette est donc *V_12*. Pour un verbe ambigu comme *sanctionner*, appartenant aux tables *6* et *12*, l'étiquette induite est *V_6_12*.

Ensuite, dans le cadre des verbes, nous avons effectué des variantes de l'expérience précédente en prenant en compte la hiérarchie des tables des verbes. Cette hiérarchie permet d'obtenir un jeu d'étiquettes de taille moins conséquente selon le niveau de regroupement utilisé. Les classes ajoutées aux étiquettes morpho-syntaxiques dépendent maintenant du verbe ainsi que du niveau spécifique dans la hiérarchie. Par exemple, le verbe *sanctionner*, ayant pour numéros de table *6* et *12*, a pour étiquette *V_QTD2* aux niveaux 1 et 2, puis *V_TD2* au dernier niveau. Dans le cas où le verbe est ambigu, le suffixe contient toutes les classes de la hiérarchie dont les numéros de table du verbe font partie. Cette expérience sera appelée *AnnotVerbes* par la suite.

En ce qui concerne les noms prédicatifs, nous ne disposons pas encore de hiérarchie des tables. Nous avons donc testé deux méthodes. Une première méthode très simple, appelée *AnnotIN*, consiste à ajouter un suffixe *IN* à l'étiquette d'un nom si ce nom est dans le lexique syntaxique, et donc s'il s'agit d'un nom prédicatif. La deuxième méthode, appelée *AnnotNoms*, consiste à créer une hiérarchie des tables des noms à partir de la table des tables des noms prédicatifs. Cette hiérarchie est faite en fonction du nombre d'arguments maximum que peut prendre un nom d'une table d'après les propriétés définitoires spécifiées pour cette table dans la table des tables. Nous disposons donc d'un seul niveau hiérarchique. Par exemple, les noms de la table *N_aa* peuvent prendre 2 arguments au maximum alors que ceux de la table *N_an04* ne peuvent en prendre qu'un seul. Les caractéristiques de chaque niveau sont indiquées dans le tableau 2[6] (le niveau 0 représente l'ensemble des tables présentes dans le Lexique-Grammaire).

---

[6] Le nombre de noms non ambigus est de 6126 pour 6175 entrées.

| Niveau | #classes | #noms | #entrées | AVG_1 | AVG_2 |
|---|---|---|---|---|---|
| 0 | 76 | 8531 | 12351 | 162 | 1.43 |
| 1 | 3 | 8531 | 12351 | 3413 | 1.2 |

TAB. 2: Caractéristiques de la hiérarchie des tables des noms

## 5. EXPÉRIENCES ET ÉVALUATIONS

Pour nos expérimentations, nous avons utilisé le corpus arboré du français, le French Treebank (appelé FTB par la suite) (Abeillé et al., 2003), contenant 20860 phrases et 540648 mots issus du journal *Le Monde* (version de 2004). Ce corpus étant de petite taille, nous avons effectué nos évaluations selon la méthode dite de validation croisée. Cette méthode consiste à découper le corpus en *p* parties égales puis à effectuer l'apprentissage sur *p-1* parties et les évaluations sur la partie restante. On peut itérer *p* fois ce processus. Cela permet donc de calculer un score moyen sur un échantillon aussi grand que le corpus initial. Dans notre cas, nous avons fixé le paramètre *p* à 10. De plus, nous avons appliqué les mêmes prétraitements sur les étiquettes morpho-syntaxiques que dans (Crabbé & Candito, 2008). C'est à dire que les étiquettes morpho-syntaxiques tiennent compte de l'annotation morphologique riche du FTB (mode des verbes, clitiques,...), ce qui conduit à obtenir un jeu de 28 étiquettes distinctes[7]. Les mots composés ont été fusionnés afin d'obtenir un unique token.

Dans les expériences qui suivent, nous allons tester l'impact de la modification du jeu d'étiquettes du corpus d'apprentissage, à savoir l'ajout des informations issues du Lexique-Grammaire décrites dans la section 4. Les résultats des évaluations sur les corpus d'évaluation sont reportés en utilisant le protocole standard PARSEVAL (Black et al., 1991) pour des phrases de taille inférieure à 40 mots. Le score de F-mesure tient compte du parenthésage et également des catégories des noeuds (en tenant compte des noeuds de ponctuation). Pour chaque expérience, nous avons indiqué les résultats *Baseline*, à savoir les résultats de BKY entraîné sur le corpus arboré original (sans annotations issues du Lexique-Grammaire). Nous avons également indiqué le pourcentage de verbes ou de noms prédicatifs annotés distincts sur la totalité du corpus pour chaque méthode d'annotation[8]. La taille du jeu d'étiquettes du corpus selon les différentes méthodes d'annotation est précisée par la colonne *Tagset* des tableaux.

### 5.1. ANNOTATION DES ÉTIQUETTES VERBALES

Nous avons tout d'abord effectué les expériences sur les verbes décrites dans la section 4, à savoir *AnnotTable* et *AnnotVerbes*. Les résultats des expériences sont montrés dans le tableau 3. Dans le cadre de l'expérience *AnnotVerbes*, nous avons fait varier deux paramètres, à savoir *Niv.* (*Niveau*) qui indique le niveau de la hiérarchie utilisé et *Amb.* (*Ambiguïté*) qui indique qu'une étiquette d'un verbe est modifiée uniquement si ce verbe appartient à un nombre de classes inférieur ou égal au nombre précisé par ce paramètre.

---

[7] Il y a 6 étiquettes différentes pour les verbes et 2 pour les noms.
[8] Le corpus contient 3058 formes verbales distinctes et 17003 formes nominales distinctes.

| Méthode | Niv./Amb. | Tagset | %verbes annotés | F-mesure/Etiquetage | Gains absolus (F-mesure) |
|---|---|---|---|---|---|
| Baseline | -/- | 28 | - | 85.05/97.43 | |
| AnnotTable | -/1 | 228 | 18,6% | 84.49/97.29 | |
| AnnotVerbes | 1/1 | 89 | 21,5% | 85.06/97.46 | |
| AnnotVerbes | 2/1 | 76 | 22,5% | 85.35/97.41 | |
| AnnotVerbes | 3/1 | 47 | 33,9% | **85.39/97.49** | |
| AnnotVerbes | 2/2 | 246 | 44,7% | 84.60/97.35 | |
| AnnotVerbes | 3/2 | 75 | 55,7% | 85.20/97.48 | |

TAB. 3: Evaluation de l'impact de la modification des étiquettes verbales

Pour les verbes non ambigus, nous pouvons observer que l'expérience *AnnotTable* dégrade fortement les performances. Cela provient très probablement de la grammaire qui est trop éclatée en raison de la taille conséquente du jeu d'étiquettes. L'effet est inversé dès que l'on utilise les niveaux de la hiérarchie des tables (niveaux 2 et 3 seulement). Les conséquences de la hiérarchisation des tables sont l'augmentation du nombre de verbes annotés comme non ambigus et la réduction de la taille du jeu d'étiquettes. Pour les niveaux 2 et 3, 6 des 10 corpus d'évaluation obtiennent un gain absolu positif situé entre +0,2% et +1%. En revanche, la prise en compte des verbes ambigus ne permet pas d'améliorer les performances (résultats montrés uniquement pour les niveaux 2 et 3 avec ambiguïté maximale de 2). La raison pourrait être identique à celle énoncée pour l'expérience *AnnotTable*, à savoir la taille conséquente du jeu d'étiquettes.

## 5.2. ANNOTATION DES ÉTIQUETTES DES NOMS PRÉDICATIFS

Pour les noms prédicatifs, nous avons effectué successivement les expériences *AnnotTable*, *AnnotNoms* et *AnnotIN*, décrites dans la section 4. Les résultats sont précisés dans le tableau 4. De même que pour les verbes, nous avons fait varier le paramètre *Ambiguïté* pour l'expérience *AnnotNoms* (le nombre de classes maximum associées à un nom étant de 3).

| Méthode | Amb. | Tagset | %noms annotés | F-mesure/ Etiquetage | Gains absolus (F-mesure) |
|---|---|---|---|---|---|
| Baseline | - | 28 | - | 85.05/97.43 | |
| AnnotTable | 1 | 98 | 8,6% | 85.10/97.42 | |
| AnnotNoms | 1 | 33 | 11,2% | 85.13/97.48 | |
| AnnotNoms | 2 | 38 | 16,5% | 85.16/97.47 | |
| AnnotNoms | 3 | 39 | 16,9% | 85.05/97.41 | |
| AnnotIN | - | 30 | 16,9% | **85.20/97.54** | |

TAB. 4: Evaluation de l'impact de la modification des étiquettes nominales

Les différentes méthodes d'annotation des noms n'augmentent que très peu les performances de l'analyseur. Contrairement aux verbes, la méthode *AnnotTable* ne dégrade pas les performances car il y a beaucoup moins de noms annotés (moins de 9%), d'où l'impact limité du nouveau jeu d'étiquettes. L'utilisation d'une hiérarchie simple des tables des noms, à travers l'expérience *AnnotNoms*, permet d'obtenir des gains positifs mais, ici, peu significatifs. On pourra cependant noter que 3 des 10 corpus d'évaluation ont été grandement améliorés (de +0.4 à +0.8). De plus, nous obtenons une légère amélioration en annotant certains noms ambigus. Étonnamment, la méthode qui donne les meilleurs résultats, malgré son principe très simple, est *AnnotIN*.

5.3. COMBINAISON DES ANNOTATIONS

Dans une dernière expérience, nous avons combiné les meilleures méthodes d'annotation des verbes et des noms prédicatifs, à savoir *AnnotIN* pour les noms prédicatifs et *AnnotVerbes* pour les verbes (niveau 3 sans ambiguïté). Les résultats sont indiqués dans le tableau 5.

| Méthode | F-mesure |
|---|---|
| Baseline | 85.05 |
| Combinaison | **85.32** |

TAB. 5: Evaluation de l'impact de la combinaison des méthodes d'annotation

La combinaison des annotations ne permet pas d'augmenter les gains obtenus avec la méthode *AnnotVerbes* et on observe même une légère dégradation.

## 6. CONCLUSION ET PERSPECTIVES

Les travaux préliminaires décrits dans cet article montrent qu'en ajoutant certaines informations issues d'un lexique syntaxique tel que le Lexique-Grammaire, nous sommes capable d'améliorer les performances d'un analyseur syntaxique probabiliste. Ces performances sont obtenues, principalement, grâce à la hiérarchie des tables des verbes qui permet de limiter l'ambiguïté en terme de nombre de classes associées à un verbe. Ceci a pour effet d'augmenter la couverture des verbes annotés selon le niveau de granularité utilisé. En revanche, dès que l'on intègre de l'ambiguïté, les performances subissent une dégradation. Les résultats obtenus sur les noms prédicatifs, notamment avec une hiérarchie simple des tables, sont peu significatifs mais laissent présager une certaine marge de progression avec une hiérarchie plus complexe comme celle disponible pour les verbes.

Dans un futur proche, nous tenterons de reproduire ces expériences en tenant compte des méthodes de clustering de mots introduites par (Candito & Crabbé, 2009; Candito & Seddah, 2010). Grâce à un algorithme semi-supervisé, leurs méthodes permettent de réduire la taille du lexique de la grammaire en regroupant les mots selon leurs comportements syntaxiques dans un corpus arboré. Ces méthodes pourraient donc être complémentaires à nos méthodes d'annotations. Une piste similaire à explorer pourrait consister à conserver le tagset original et à remplacer les tokens du corpus par les annotations syntaxiques générées par nos méthodes.